\documentclass{article}

\usepackage{graphicx}
\usepackage{lipsum}  
\usepackage{xcolor}
\usepackage[colorlinks = true,
            linkcolor = black,
            urlcolor  = blue,
            citecolor = black,
            anchorcolor = blue]{hyperref}
\usepackage{hang}
\usepackage{geometry}
\usepackage{titlesec}
\usepackage{indentfirst}

\usepackage{authblk}
\usepackage{comment}

\titleformat*{\section}{\LARGE\bfseries}
\titleformat*{\subsection}{\bfseries}
\usepackage{natbib}
\title{\textbf{Datasheet for the Pile}}
\author{Stella Biderman}
\author{Kieran Bicheno}
\author{Leo Gao}
\affil{EleutherAI \\ \texttt{contact@eleuther.ai}}

\begin{document}
\maketitle
\begin{abstract}
This datasheet describes the Pile, a $825$ GiB dataset of human-authored text compiled by EleutherAI for use in large-scale language modeling. The Pile is comprised of 22 different text sources, ranging from original scrapes done for this project, to text data made available by the data owners, to third-party scrapes available online.
\end{abstract}

\section{Background on the Pile}

The Pile is a massive text corpus created by EleutherAI for large-scale language modeling efforts. It is comprised of textual data from 22 sources (see below) and can be downloaded from \href{https://pile.eleuther.ai/}{the official website} as well as from a \href{https://mystic.the-eye.eu/public/AI/pile/}{community mirror}. Each source dataset is at its core a textual work, and any non-textual data (including metadata) has been removed. While still preserving their internal order, the documents from all the sources have been randomly shuffled. For further information on the Pile, see \citet{pile}.

\textbf{This document is not intended to be -- and should not be used as -- a substitute for a datasheet for the original versions of the component datasets}. While it is accurate for the text data that we derived from each component dataset, the original source dataset may have other properties. This document is intended to inform people interested in using the Pile for natural language processing. People interested in using the original datasets should contact the data owners for information about the properties of the original data.

It is not always the case that the answer to the questions below are known with certainty. For example, while we have no reason to believe that personal identifying information (PII) is contained in most of the subsets of our dataset, it is always possible that someone wrote down PII in a document and uploaded it to arXiv. Due to the sheer scale of the data, it is impractical to systematically search through every text to validate that it is what it purports to be. We have endeavored to answer the questions below as best we can, and to be open and honest about the limitations of the accuracy of this document. Anyone who engages in research on or with the Pile is welcome to contact us to have their findings added to this document. Similarly, we welcome all comments, suggestions, or corrections. 

\parindent=0pt
\leftskip=15pt

\subsection*{Datasets contained in the Pile:}

\textbf{Pile-CC:} The Pile-CC dataset is a sample from the Common Crawl WARCs that has been converted to text using jusText \citep{justext}.

\textbf{PubMed Central:} The PubMed Central dataset is a subset of the PubMed online  repository  for  biomedical  articles  run  by the United States  of America’s National Center for Biotechnology Information (NCBI).

\textbf{Books3:} The Books3 component is a dataset of books derived from a copy of the contents of the Bibliotik private tracker made available by The Eye \citep{Bibliotik}.

\textbf{arXiv:} The arXiv component is a subset of the ArXiv preprint repository for research papers that has operated since 1991.

\textbf{Github:} The GitHub component is an EleutherAI scrape of GitHub, a large dataset of open source code repositories.

\textbf{OpenWebText2:} The OpenWebText2 component is a web scrape dataset produced by EleutherAI and inspired by WebText \citep{GPT2} and OpenWebTextCorpus \citep{OpenWeb}.

\textbf{FreeLaw:} The Free Law Project is US registered non-profit that provide access to millions of legal opinions and analytical tools for academic studies in the legal realm. 

\textbf{Wikipedia (en):} The Wikipedia (en) dataset is taken from the Wikipedia site as a standard source of high-quality text for language modeling. 

\textbf{StackExchange:} The StackExchange dataset is a dump of anonymized user-contributed content on the Stack Exchange network, a popular collection of websites centered around user-contributed questions and answers.

\textbf{USPTO Backgrounds:} The USPTO Backgrounds dataset is a set of background sections from patents granted by the United States Patent and Trademark Office, derived from its published bulk archives\footnote{\url{https://bulkdata.USPTO.gov/}}.

\textbf{PubMed Abstracts:} The PubMed Abstracts dataset comprises the abstracts of 30 million publications in the PubMed online repository for biomedical articles. 

\textbf{Project Gutenberg (PG-19):} The Project Gutenberg dataset is a corpus of high-quality, classic literature.

\textbf{OpenSubtitles:} The OpenSubtitles dataset is an English language dataset of subtitles from movies and television shows gathered by \citet{OpenSubtitles}. 

\textbf{DM Mathematics:} The DeepMind Mathematics dataset consists of a collection of mathematical problems such as algebra, arithmetic, calculus, number theory, and probability, formatted as natural language prompts \citep{dm-mathematics}.

\textbf{BookCorpus2:} BookCorpus2 is an expanded version of the original BookCorpus \citep{BookCorpus}, a widely used language modeling corpus consisting of books written by ``as of yet unpublished authors.''

\textbf{Ubuntu IRC:} The Ubuntu IRC dataset is derived from the publicly available chatlogs\footnote{\url{https://irclogs.ubuntu.com/}} of all Ubuntu-related channels on the Freenode IRC chat server.  

\textbf{EuroParl:} EuroParl \citep{EuroParl} is a multilingual parallel corpus originally introduced for machine translation but which has also seen use in several other fields of NLP \citep{EuroParl-example1,EuroParl-example2,EuroParl-example3}.

\textbf{YouTube Subtitles:} The YouTube Subtitles dataset is a parallel corpus of text gathered from human generated closed-captions on YouTube.

\textbf{PhilPapers:} PhilPapers\footnote{\url{https://philpapers.org/}} is a dataset of open access philosophy publications from an international database maintained by the Center for Digital Philosophy at the University of Western Ontario.

\textbf{NIH ExPORTER:} The NIH Grant abstracts provides a bulk-data repository for awarded applications through the ExPORTER\footnote{\url{https://exporter.nih.gov/}} service covering the fiscal years 1985-present. 

\textbf{HackerNews:} Hacker News\footnote{\url{https://news.ycombinator.com}} is a link aggregator operated by Y Combiner, a startup incubator and investment fund. 

\textbf{Enron Emails:} The Enron Emails dataset \citep{Enron} contains text from the contents of Enron's email servers unearthed during the investigation into that organization's accounting methods and is a valuable corpus for understanding the modality of email communications, which are typically not found in any of our other datasets. 

\section{Motivation For Dataset Creation}

\subsection*{Why was the dataset created? (e.g., was there a specific task in mind? was there a specific gap that needed to be filled?)}

\textbf{The Pile:} The Pile was created for the purposes of training large-scale language models such as \citet{GPT2}, \citet{TuringNLG}, and \citet{GPT3}. Training such models requires massive amounts of human-authored text data which, hitherto, was not available from a single source other than Pile-CC derived data. Below we discuss the motivation for creating each constituent collection. For data collections that were created specifically for the Pile, we also discuss the motivations for using the underlying data collection.\\

\textbf{Pile-CC:} The Pile-CC dataset was created to be included in the Pile. The underlying data comes from the Common Crawl, which was created to give people access to the wealth of information contained in the internet. Its creators were concerned that only data mining companies would be able to collect this data, and has the explicit aim of democratizing technology.

\textbf{PubMed Central:} The PubMed Central dataset was created to be included in the Pile. The underlying data comes from the PubMed Central database, which was created to allow public access to the results of research funded by the U.S. National Institute of Health and in compliance with the Consolidated Appropriations Act of 2008 (H.R. 2764).

\textbf{Books3:} The Books3 dataset was created as a resource for language modelling research. We included Books3 \citep{Bibliotik}, a pre-existing collection of long-form books, because books are invaluable for long-range context modeling research.

\textbf{arXiv:} The arXiv dataset was created to be included in the Pile. We included arXiv in the hopes that it will be a source of high quality text and math knowledge, and benefit potential downstream applications to research in Math, CS, Physics, and Machine Learning.

\textbf{Github:} The GitHub dataset was created to be included in the Pile. The data comes from the Github website\footnote{\url{https://github.com/}} which was created for hosting offsite Git\footnote{\url{https://git-scm.com/}} software repositories 

\textbf{OpenWebText2:} The OpenWebText2 dataset was created to be included in the Pile. We wanted a high-quality web scrape similar to WebText \citep{GPT2} and OpenWebTextCorpus \citep{OpenWeb}.

\textbf{FreeLaw:} The FreeLaw dataset was created to be included in the Pile. The underlying data comes from the Free Law Project. According to the Free Law Project website, they have several complimentary purposes for collecting and publishing the data:
\begin{quote}
    to provide free, public, and permanent access to primary legal materials on the Internet for educational, charitable, and scientific purposes to the benefit of the general public and the public interest; to develop, implement, and provide public access to technologies useful for legal research; to create an open ecosystem for legal research and materials; to support academic research on related technologies, corpora, and legal systems; and to carry on other charitable activities associated with these purposes, including, but not limited to, publications, meetings, conferences, trainings, educational seminars, and the issuance of grants and other financial support to educational institutions, foundations, and other organizations exclusively for educational, charitable, and scientific purposes as allowed by law.
\end{quote}

\textbf{Wikipedia (en):} The Wikipedia (en) dataset was created as a standard source of high-quality text for language modeling and a source of information in question-answer models.

\textbf{StackExchange:} The StackExchange dataset was created to be included in the Pile. The underlying data comes from a dump of StackExchange questions and answers which was created to provide people with the ability to analyze the data contained in the StackExchange network.

\textbf{USPTO Backgrounds:} The USPTO dataset was created to be included in the Pile. The underlying data comes from the US Patent and Trademark Office's website, and the data sets were created ``[t]o advance research on matters relevant to intellectual property, entrepreneurship, and innovation,'' to ``facilitate economic research on patents and trademarks'' and to ``support White House policy that champions transparency and access to government data under the "data.gov" umbrella of initiatives.''

\textbf{PubMed Abstracts:} The PubMed Abstracts dataset was created to be included in the Pile. The underlying data comes from the PubMed publication index which was created as part of fulfilling the U.S. National Institute of Health's ``charge of maintaining a record of biomedical  research  of  grants,  publications,  and  other scholarly works'' and providing public access to taxpayer funded research.

\textbf{Project Gutenberg (PG-19):} The PG-19 dataset was created by DeepMind from the Gutenberg Project, and is included as a source of high-quality classic literature and the source of many English idioms useful in downstream NLP projects.

\textbf{OpenSubtitles:} The OpenSubtitles dataset was create to serve as a ``prime resource for the compilation of parallel corpora.''\citep{OpenSubtitles}.

\textbf{DM Mathematics:} The DM Mathematics dataset was created by DeepMind to aid investigation of the ability of neural networks to learn to solve arithmetical problems.

\textbf{BookCorpus2:} The BookCorpus2 dataset was created for the Pile to aid in natural language processing research.

\textbf{Ubuntu IRC:} The Ubuntu IRC corpus was created as a ``unique resource for research into building dialogue managers based on neural language models that can make use of large amounts of unlabeled data'' \citep{lowe2016ubuntu}. 

\textbf{EuroParl:} The EuroParl dataset was created to aid statistical machine translation research.

\textbf{YouTube Subtitles:} The YouTube Subtitles dataset was created to be included in the Pile.

\textbf{PhilPapers:} The PhilPapers dataset was created to be included in the Pile.

\textbf{NIH ExPORTER:} The NIH ExPORTER database was created to provide public access to biomedical research.

\textbf{HackerNews:} The HackerNews dataset was created to be included in the Pile.

\textbf{Enron Emails:} The Enron Emails dataset was created to aid investigation into the unethical accounting practices of Enron and the circumstances surrounding its collapse. The corpus is not the email data itself but text derived from it for what has become routine research into email and email users' behaviour.

\subsection*{Has the dataset been used already? If so, where are the results so others can compare (e.g., links to published papers)?}

\textbf{The Pile:} The Pile has been used as a trainin dataset for a variety of large language models including GPT-Neo \citep{black2021gpt}, GPT-J \citep{wang2021gpt}, Jurassic-1 \citep{lieber2021jurassic}, Wu Dao \citep{WuDao}, and GPT-NeoX \citep{gpt-neox}. Other models, trained using custom subsets of the Pile, are listed below by subset. A number of papers study the properties of models trained on the Pile, including \citet{mitchell2021fast,peyrard2021invariant,matiana2021cut,mukherjee2021neural,magee2021intersectional}, and \citet{lee2021deduplicating}. \\

\textbf{Pile-CC:} Papers that train models on datasets that include the Pile-CC subset of the Pile include \citet{luo2021analyzing,Megatron-Turing,askell2021general}. While this data has likely been used by other researchers for various purposes, we are unaware of any uses that would be directly comparable.

\textbf{PubMed Central:} None

\textbf{Books3:} Papers that train models on datasets that include the Books3 subset of the Pile include \citet{luo2021analyzing,wang2021phrase,Megatron-Turing}. While this data has likely been used by other researchers for various purposes, we are unaware of any uses that would be directly comparable.

\textbf{arXiv:} Papers that train models on datasets that include the arXiv subset of the Pile include \citet{askell2021general,Megatron-Turing}. While this data has likely been used by other researchers for various purposes, we are unaware of any uses that would be directly comparable.

\textbf{Github:} Papers that train models on datasets that include the GitHub subset of the Pile include \citet{askell2021general,Megatron-Turing}. While this data has likely been used by other researchers for various purposes, we are unaware of any uses that would be directly comparable.

\textbf{OpenWebText2:} Papers that train models on datasets that include the OpenWebText2 subset of the Pile include \citet{luo2021analyzing,Megatron-Turing}

\textbf{FreeLaw:} Papers that train models on datasets that include the FreeLaw subset of the Pile include \citet{askell2021general}. While this data has likely been used by other researchers for various purposes, we are unaware of any uses that would be directly comparable.

\textbf{Wikipedia (en):} Papers that train models on datasets that include the Wikipedia (en) subset of the Pile include \citet{luo2021analyzing,Megatron-Turing}. While this data has likely been used by other researchers for various purposes, the only ones that we are aware of that is directly comparable are \citet{GPT2,GPT3}.

\textbf{StackExchange:} Papers that train models on datasets that include the StackExchange subset of the Pile include \citet{Megatron-Turing}. While this data has likely been used by other researchers for various purposes, we are unaware of any uses that would be directly comparable.

\textbf{USPTO Backgrounds:} Papers that train models on datasets that include the USPTO Backgrounds subset of the Pile include \citet{askell2021general}. While this data has likely been used by other researchers for various purposes, we are unaware of any uses that would be directly comparable.

\textbf{PubMed Abstracts:} Papers that train models on datasets that include the PubMed Abstracts subset of the Pile include \citet{Megatron-Turing}. While this data has likely been used by other researchers for various purposes, we are unaware of any uses that would be directly comparable.

\textbf{Project Gutenberg (PG-19):} Project Gutenberg 1919 has been used extensively, including by \citep{PG19-1,PG19-2,PG19-3,PG19-4,PG19-5}. Papers that train models on datasets that include the PG-19 subset of the Pile include \citet{Megatron-Turing}.

\textbf{OpenSubtitles:} OpenSubtitles has been used extensively, including by \citep{OSub-1,OSub-2,OSub-3,OSub-4,OSub-5}. Papers that train models on datasets that include the OpenSubstitles subset of the Pile include \citet{luo2021analyzing,askell2021general}

\textbf{DM Mathematics:} DM Mathematics has been used extensively, including by \citep{DMM-1,DMM-2,DMM-3,DMM-4,DMM-5}.

\textbf{BookCorpus2:} The BookCorpus dataset that BookCorpus2 is based on has been used extensively, including by \citep{BkCo-1,BkCo-2,BkCo-3,BkCo-4,BkCo-5}. BookCorpus2 was studied by \citet{bandy2021addressing}. Papers that train models on datasets that include the Pile-CC subset of the Pile include \citet{Megatron-Turing,askell2021general}

\textbf{Ubuntu IRC:} Papers that train models on datasets that include the Pile-CC subset of the Pile include \citet{askell2021general}. While this data has likely been used by other researchers for various purposes, we are unaware of any uses that would be directly comparable. 

\textbf{EuroParl:} EuroParl has been extensively used, including by \citep{Euro-1,Euro-2,Euro-3,Euro-4,Euro-5}.

\textbf{YouTube Subtitles:} While this data has likely been used by other researchers for various purposes, we are unaware of any uses that would be directly comparable.

\textbf{PhilPapers:} While this data has likely been used by other researchers for various purposes, we are unaware of any uses that would be directly comparable.

\textbf{NIH ExPORTER:} Papers that train models on datasets that include the NIH ExPORTER subset of the Pile include \citet{Megatron-Turing}. While this data has likely been used by other researchers for various purposes, we are unaware of any uses that would be directly comparable. 

\textbf{HackerNews:} While this data has likely been used by other researchers for various purposes, we are unaware of any uses that would be directly comparable.

\textbf{Enron Emails:} The Enron Emails have been used extensively, including by \citep{Enrn-1, Enrn-2, Enrn-3, Enrn-4, Enrn-5}.

\subsection*{What (other) tasks could the dataset be used for?}

\textbf{The Pile:} Answers differ by the subset. For each component document, see below.\\

\textbf{Pile-CC:} The Pile-CC dataset could be used for statistical analysis and comparisons of online dialects in NLP, or structural analysis for non-ML research disciplines.

\textbf{PubMed Central:} The PubMed dataset will allow researchers to test the efficacy of a large-scale general model that includes technical medical information in communication, explanation, and data-mining fields. It could also be used to ensure the legal and regulatory instruments that mandate the dataset's creation are being followed.

\textbf{Books3:} The Books3 dataset could be used as a batch of templates for copyright enforcement checks, as well as allowing models based on this corpus to be used by researchers in the fields of literature and creative arts.

\textbf{arXiv:} The arXiv dataset could be mined for potential answers to significant academic pursuits, assuming many submissions remain broadly unread. Including a deep set of technical papers in the corpora and its downstream models will give researchers a tool with both general and specific understanding of their domain.

\textbf{Github:} Including a large base of code in a structured way gives downstream models the potential for aiding researchers in creating code for their theoretical work without needing to know how to code themselves.

\textbf{OpenWebText2:} The OpenWebText2 dataset's inclusion allows downstream research tools access to insights not formally captured in a traditional setting like arXiv or PubMed.

\textbf{FreeLaw:} The FreeLaw dataset could be used as a rolling track of the linguistic accessibility of judgments and legal arguments. AI digital assistants or researcher tools created with this corpus would also have an in-built lexicon for understanding legal requirements, such as their own Terms of Use.

\textbf{Wikipedia (en):} The Wikipedia dataset gives a powerful question-and-answer capability for researchers in a broad set of domains and a curated source of fact-checking capacity.

\textbf{StackExchange:} The StackExchange dataset, especially when combined with the GitHub corpus, gives researchers the potential ability to create self-diagnosing and self-repairing models. This also gives the Pile a basis for creating technical support chatbots.

\textbf{USPTO Backgrounds:} The USPTO dataset's inclusion gives researchers a basis for downstream problem-solving models thanks to the large body of problem-based situational awareness built into The Pile.

\textbf{PubMed Abstracts:} The PubMed Abstracts dataset gives researchers, especially in the medical field, a strong basis for building models for deep data-mining and pattern detection in collections of data too large to be dealt with by humans.

\textbf{Project Gutenberg (PG-19):} Training on the PG-19 dataset gives downstream models access to a large number of idiomatic language. The dataset also provides a basis for researchers to track linguistic changes in the English language over time.

\textbf{OpenSubtitles:} The OpenSubtitles data set gives researchers access to better-structured common-language understanding in their downstream projects. 

\textbf{DM Mathematics:} The DeepMind Mathematics data set would give researchers in the educational space a platform for testing smaller corpora on their ability to handle mathematical Q\&A capabilities against those given by The Pile.

\textbf{BookCorpus2:} BookCorpus2 has a secondary role for researchers as a benchmark for the efficacy of models based on more focused corpora, as BookCorpus2 adds depth of training to the otherwise wide capabilities of The Pile.

\textbf{Ubuntu IRC:} The Ubuntu IRC data set could be used for building chatbots or training self-diagnostic models for computers.

\textbf{EuroParl:} The EuroParl texts add a layer of multi-lingual benchmarking for researchers testing results between specific modals and general models such as those based on The Pile.

\textbf{YouTube Subtitles:} The YouTube subtitles dataset also gives those researchers using the Pile a multi-lingual capability among the predominantly English remainder of the corpus.

\textbf{PhilPapers:} The PhilPapers dataset could be mined for an exhaustive map of intellectual domains by researchers involved in ontology-based projects.

\textbf{NIH ExPORTER:} The NIH ExPORTER dataset gives downstream models a framework by which researchers can add an understanding of government and NGO processes to their models.

\textbf{HackerNews:} HackerNews provides The Pile and downstream models a non-technical bridge between domain specific lexicons and everyday speech in a way that researchers can exploit in projects such as data assistants.

\textbf{Enron Emails:} The Enron emails could be used to study the activities of human agents during a large-system collapse and provide researchers using downstream tools to interface more naturally with email-based projects.

\subsection*{Who funded the creation of the dataset?}

\textbf{The Pile:} The Pile was created by EleutherAI. This dataset was created by individuals working on their own time without funding. All components that required processing beyond their orignal form were also processed by EleutherAI members on their own time without any funding. \\

\textbf{Pile-CC:} The data is sourced from Common Crawl, a non-profit 501(c)(3) organization founded by Gil Elbaz. The data from Common Crawl was processed by EleutherAI into Pile-CC.

\textbf{PubMed Central:} The U.S. National Center for Biotechnology Information (NCBI), a branch of the National Institutes of Health (NIH).

\textbf{Books3:} Books3 was created by Shawn Presser, an independent ML developer.

\textbf{arXiv:} EleutherAI scraped arXiv ourselves.

\textbf{Github:} EleutherAI scraped Github ourselves.

\textbf{OpenWebText2:} EleutherAI created OpenWebText2 ourselves.

\textbf{FreeLaw:} The Free Law Project, a 501(c)(3) non-profit founded by  Michael Lissner and Brian Carver.

\textbf{Wikipedia (en):} The Wikimedia Foundation, a 501(c)(3) organization founded by Jimmy Wales, was the source of the data and DeepMind, an Alphabet subsidiary, created the dataset.

\textbf{StackExchange:} EleutherAI scraped the StackExchange Network ourselves.

\textbf{USPTO Backgrounds:} The U.S. Office of the Chief Economist.

\textbf{PubMed Abstracts:} The U.S. National Institute of Health.

\textbf{Project Gutenberg (PG-19):} Project Gutenberg has received funding from a variety of sources, including the University of Illinois, Carnegie Melon University, and University of North Carolina at Chapel Hill. We believe that its costs are currently paid for by UNC Chapel Hill, but are not certain. The form of the dataset we used was created by DeepMind, an Alphabet subsidiary \citep{PG19}.

\textbf{OpenSubtitles:} The site is largely, if not entirely, volunteer-created but paid for by members paying for `VIP Membership'.

\textbf{DM Mathematics:} DeepMind, an artificial intelligence company and research laboratory owned by Alphabet Inc. (the parent company of Google inc.)

\textbf{BookCorpus2:} The underlying data was created by individual ``as of yet unpublished authors". EleutherAI processed the BookCorpus2 dataset ourselves.

\textbf{Ubuntu IRC:} Ubuntu itself is funded by Canonical Ltd., a private company founded by Mark Shuttleworth, and the conversations that constitute the dataset were on the Freenode IRC server, funded through donations and volunteers.

\textbf{EuroParl:} We believe that this dataset was funded the School of Informatics of the University of Edinburgh, Scotland, but were unable to confirm this fact. For further information, contact \citet{groves2006hybridity}.

\textbf{YouTube Subtitles:} EleutherAI scraped YouTube for subtitles ourselves.

\textbf{PhilPapers:} EleutherAI scraped PhilPapers ourselves. The underlying data collection was funded by the Centre for Digital Philosophy at the University of Western Ontario.

\textbf{NIH ExPORTER:} EleutherAI scraped PhilPapers ourselves. The underlying data collection was funded by the U.S. National Institute of Health and other agencies of the U.S. Department of Health and Human Services (ACF, AHRQ, CDC, HRSA, FDA), and the VA.

\textbf{HackerNews:} EleutherAI scraped the Hacker News website ourselves. Hacker News is funded by YCombinator.

\textbf{Enron Emails:} The U.S. Federal Energy Regulatory Commission (FERC).

\section{Dataset Composition}

\subsection*{What are the instances?(that is, examples; e.g., documents, images, people, countries) Are there multiple types of instances? (e.g., movies, users, ratings; people, interactions between them; nodes, edges)}

\textbf{The Pile:} Instances of the dataset are textual documents of a variety of contents. The instances come from the categories described below.\\

\textbf{Pile-CC:} Instances are webpages.

\textbf{PubMed Central:} Instances are academic medical papers.

\textbf{Books3:} Instances are published books.

\textbf{arXiv:} Instances are preprints of academic papers, primarily in mathematics, computer science, physics, and statistics.

\textbf{Github:} Instances are code files.

\textbf{OpenWebText2:} Instances are webpages.

\textbf{FreeLaw:} Instances are legal documents.

\textbf{Wikipedia (en):} Instances are pages of Wikipedia (en).

\textbf{StackExchange:} Instances are questions posted on StackExchange, along with highly upvoted answers.

\textbf{USPTO Backgrounds:} Instances are abstracts of US Patent applications.

\textbf{PubMed Abstracts:} Instances are abstracts of papers in the PubMed archive, primarily published medical papers.

\textbf{Project Gutenberg (PG-19):} Instances are books published prior to 1919.

\textbf{OpenSubtitles:} Instances are the subtitles for a movie.

\textbf{DM Mathematics:} Instances are groups of related mathematics problems.

\textbf{BookCorpus2:} Instances are books.

\textbf{Ubuntu IRC:} Instances are chatlogs, chunked by week.

\textbf{EuroParl:} Instances are minutes from meetings of the European Parliament. A large proportion of the data contains the same text repeated in different languages. Copies of the same text in different languages are separate documents.

\textbf{YouTube Subtitles:} Instances are subtitles for YouTube videos.

\textbf{PhilPapers:} Instances are preprints of academic papers, primarily in philosophy.

\textbf{NIH ExPORTER:} Instances are medical academic papers.

\textbf{HackerNews:} Instances are conversation threads from the Hacker News Network.

\textbf{Enron Emails:} Instances are emails between employees of Enron from January 6, 1998 until February 4, 2004.

\subsection*{How many instances are there in total (of each type, if appropriate)?}
The number of documents reported are after deduplication. Note that the Pile has had its datasets weighted. The sizes reported here are the raw sizes. For the effective sizes in the Pile, see Figure 1 of \citet{pile}.

\textbf{The Pile:} $211,043,181$ documents (unweighted), totaling $825.18$ GiB.\\

\textbf{Pile-CC:} $54,953,117$ documents, totaling $227.12$ GiB.

\textbf{PubMed Central:} $3,098,931$ documents, totaling $90.27$ GiB.

\textbf{Books3:} $196,640$ documents, totaling $100.96$ GiB.

\textbf{arXiv:} $1,264,405$ documents, totaling $56.21$ GiB.

\textbf{Github:} $19,021,454$ documents, totaling $95.16$ GiB.

\textbf{OpenWebText2:} $17,103,059$ documents, totaling $62.77$ GiB.

\textbf{FreeLaw:} $3,562,015$ documents, totaling $51.15$ GiB.

\textbf{Wikipedia (en):} $6,033,151$, totaling $6.38$ GiB

\textbf{StackExchange:} $15,622,475$ documents, totaling $32.20$ GiB.

\textbf{USPTO Backgrounds:} $5,883,037$ documents, totaling $22.90$ GiB.

\textbf{PubMed Abstracts:} $15,518,009$ documents, totaling $19.26$ GiB.

\textbf{Project Gutenberg (PG-19):} $28,602$ documents, totaling $10.88$ GiB.

\textbf{OpenSubtitles:} $446,612$ documents, totaling $12.98$ GiB.

\textbf{DM Mathematics:} $1,014,997$ documents, totaling $7.75$ GiB.

\textbf{BookCorpus2:} $17,868$ documents, totaling $6.30$ GiB.

\textbf{Ubuntu IRC:} $10,605$ documents, totaling $5.52$ GiB.

\textbf{EuroParl:} $69,814$ documents, totaling $4.59$ GiB.

\textbf{YouTube Subtitles:} $173,651$ documents, totaling $3.73$ GiB.

\textbf{PhilPapers:} $33,990$ documents, totaling $2.38$ GiB.

\textbf{NIH ExPORTER:} $939,668$ documents, totaling $1.89$ GiB.

\textbf{HackerNews:} $831,198$ documents, totaling $3.90$ GiB.

\textbf{Enron Emails:} $517,401$ documents, totaling $0.88$ GiB.

\subsection*{What data does each instance consist of? “Raw” data (e.g., unprocessed text or images)? Features/attributes? Is there a label/target associated with instances? If the instances related to people, are subpopulations identified (e.g., by age, gender, etc.) and what is their distribution?}
\textbf{The Pile:} Instances are text files, processed for readability and to remove autogenerated text, garbage generated during parsing, and code fragments picked up from websites. Some of this junk undoubtedly made it through our processing and is embedded in the text files.

\subsection*{Is there a label or target associated with each instance? If so, please provide a description.}
\textbf{The Pile:} No.

\subsection*{Is any information missing from individual instances? If so, please provide a description, explaining why this information is missing (e.g., because it was unavailable). This does not include intentionally removed information, but might include, e.g., redacted text.}
\textbf{The Pile:} Not as far as we are aware.

\subsection*{Are relationships between individual instances made explicit (e.g., users’ movie ratings, social network links)? If so, please describe how these relationships are made explicit.}
\textbf{The Pile:} Not as far as we are aware.

\subsection*{Does the dataset contain all possible instances or is it a sample (not necessarily random) of instances from a larger set? If the dataset is a sample, then what is the larger set? Is the sample representative of the larger set (e.g., geographic coverage)? If so, please describe how this representativeness was validated/verified. If it is not representative of the larger set, please describe why not (e.g., to cover a more diverse range of instances, because instances were withheld or unavailable).}
\textbf{The Pile:} Answers differ by the subset. For each component document, see below. Many of the underlying datasets grow over time, in which case the date of collection is included. A more complete explanation can be found in \citet{pile}, Appendix C. For all components, we have no idea how representative it is of the relevant whole reference class.\\

\textbf{Pile-CC:} A tiny fraction of the entire Common Crawl was included, chosen arbitrarily and heavily filtered as detailed in \citet{pile}.

\textbf{PubMed Central:} All articles in PubMed Central (as of June 2020) are contained in the Pile.

\textbf{Books3:} The entire Books3 dataset is contained in the Pile.

\textbf{arXiv:} We downloaded the \TeX\ sources of all papers on arXiv up to the July 2020 dump (the last file included in our data is {\tt arXiv\_src\_2007\_068.tar}) via arXiv's S3 Bulk Source File Access\footnote{\url{https://arxiv.org/help/bulk_data_s3}}, and used {\tt pandoc 1.19.2.4} to convert these source files to Markdown, discarding any papers which had errors during the conversion process.

\textbf{Github:} The underlying data taken from Github is filtered for only small repositories and only files smaller than 100KB, and subsampled to 95 GiB from the original 600 GiB of files matching the previous criteria. For full details, see \citet{pile}.

\textbf{OpenWebText2:} To produce the dataset, URLs and their associated metadata were first extracted from all Reddit submissions up to April 2020. URLs were deduplicated, with each unique URL featuring a list of associated submissions metadata, and an aggregate score. URLs with an aggregate score of less then 3 were removed. The links were then scraped and processed with Newspaper scraper. Deduplication was performed at the document level using in memory MinHashLSH through the DataSketch library.

\textbf{FreeLaw:} We retained the subset consisting of court opinions and excluded dockets, people, retention votes, and citation data as they were broadly administrative. Court opinions consisted of either plaintext or html. For opinions provided in html, all formatting was discarded and the raw text with extracted using \texttt{BeautifulSoup}.

\textbf{Wikipedia (en):} The entire  \texttt{wikipedia/20200301.en} dataset\footnote{\url{https://www.tensorflow.org/datasets/catalog/wikipedia\#wikipedia20200301en}} is included in the Pile.

\textbf{StackExchange:} To construct the dataset, we download and parse every Stack Exchange database dump as of July 1, 2020 to plaintext files. We opt to extract the top three answers with at least three upvotes, discarding all other responses. We only include the plain text question and response and do not incorporate any metadata.

\textbf{USPTO Backgrounds:} The United States Patent and Trademark Office (USPTO) has published bulk archives of the full text of all patents granted in the US from 1976 to September 2020. From these archives, we extract the Background sections, along with key grant-specific metadata, such as the inventor, assignee, and classification information.

\textbf{PubMed Abstracts:} About one-third of the articles in the dataset were missing or contained a malformed title or abstract and were excluded. Additionally, PubMed Central contains full-text resources to many recent publications; any publications which already appear in PMC are excluded from this set.

\textbf{Project Gutenberg (PG-19):} The entire Project Gutenberg 1919 dataset is contained in the Pile.

\textbf{OpenSubtitles:} All datapoints tagged as English by \citet{OpenSubtitles}. We discarded any provided metadata.

\textbf{DM Mathematics:} The entire DM Mathematics dataset is contained in the Pile.

\textbf{BookCorpus2:} The original BookCorpus consists of 11,038 books. However, due to issues with availability of the original BookCorpus, as well as the possibility of collecting a larger version, we decided to collect our own version of BookCorpus using a similar methodology as \citet{BookCorpusCode}. Our version of BookCorpus contains 17,868 books instead, due to the expanding nature of the underyling data.

\textbf{Ubuntu IRC:} We processed all logs from July 5, 2004 through September 1, 2020. All system messages, such as joins, disconnects, nick changes, etc. were discarded, but actions (i.e using {\tt /me}) were kept.

\textbf{EuroParl:} We download the data in bulk from \footnote{\url{ http://www.statmt.org/europarl/}}. We remove all basic tag information and only retain the name of each document as a title. For example, {\tt <SPEAKER ID=77 LANGUAGE="NL" NAME="Pronk">} becomes {\tt Pronk}, and then extract the body of each document, discarding those that are shorter than 200 characters.

\textbf{YouTube Subtitles:} The entire YouTube Subtitles dataset is contained in the Pile.

\textbf{PhilPapers:} The entire PhilPapers dataset (as of [date]) is contained in the Pile.

\textbf{NIH ExPORTER:} The NIH provides a bulk-data repository for awarded applications through the ExPORTER service covering the fiscal years 1985--present. These data come from the NIH, but also other other Health and Human Services agencies (ACF, AHRQ, CDC, HRSA, FDA), and the VA. Additionally, the NIH provides a legacy data format named CRISP for awarded applications during the fiscal years 1970--2009. We merged both the ExPORTER and CRISP data to form a consolidated dataset of awarded applications. Entries were deduplicated based off their application ID, and excluded if their abstract text was missing or too short. Small grants, especially administrative ones, consisted solely of short boilerplate. For this reason, we further deduplicated on abstract text. All grants types were considered, including new applications (Application Type Code 1) and renewals (Application Type Code 2) as the text differed enough to provide novel input. The text was then minimally parsed to remove administrative boilerplate, (ex. most old awards contain some variation of ``description: (provided by applicant)"). In total, there were 939,668 grant application abstracts added.

\textbf{HackerNews:} The entire HackerNews dataset (as of [date]) is contained in the Pile.

\textbf{Enron Emails:} The entire Enron Emails dataset is contained in the Pile.

\subsection*{Are there recommended data splits (e.g., training, development/validation, testing)? If so, please provide a description of these splits, explaining the rationale behind them.}
\textbf{The Pile:} Yes. The Pile comes with recommended train/validation/test splits. The validation and test datasets are $0.1\%$ of the data each, or about 1.4 GiB. They were chosen at random from the entire dataset.

\subsection*{Are there any errors, sources of noise, or redundancies in the dataset? If so, please provide a description.}
\textbf{The Pile:} In the Pile, some components are deliberately upsampled (see \citet{pile}). In terms of the constituent datasets, we have sought to deduplicate instances in Pile-CC and OpenWebText2, as detailed in \citet{pile}. All datasets were machine processed, and therefore likely to contain unknown noise.

\subsection*{Is the dataset self-contained, or does it link to or otherwise rely on external resources (e.g., websites, tweets, other datasets)? If it links to or relies on external resources, a) are there guarantees that they will exist, and remain constant, over time; b) are there official archival versions of the complete dataset (i.e., including the external resources as they existed at the time the dataset was created); c) are there any restrictions (e.g., licenses, fees) associated with any of the external resources that might apply to a future user? Please provide descriptions of all external resources and any restrictions associated with them, as well as links or other access points, as appropriate.}
\textbf{The Pile:} All datasets in the Pile are self-contained.

\section{Collection Process} 

\subsection*{What mechanisms or procedures were used to collect the data (e.g., hardware apparatus or sensor, manual human curation, software program, software API)? How were these mechanisms or procedures validated?}
\textbf{The Pile:} Detailed information about how each dataset was collected and processed can be found in the appendix of \citet{pile}. The actual code used can be found on \href{https://github.com/EleutherAI/The-Pile}{GitHub}.

\subsection*{How was the data associated with each instance acquired? Was the data directly observable (e.g., raw text, movie ratings), reported by subjects (e.g., survey responses), or indirectly inferred/derived from other data (e.g., part-of-speech tags, model-based guesses for age or language)? If data was reported by subjects or indirectly inferred/derived from other data, was the data validated/verified? If so, please describe how.}
\textbf{The Pile:} Answers differ by the subset. For each component document, see below.\\

\textbf{Pile-CC:} Data in the Pile-CC dataset were scraped from websites by the Common Craw and then downloaded directly from the Common Craw by EleutherAI.

\textbf{PubMed Central:} Papers in the PubMed Central dataset were uploaded to PubMed Central by the authors and then downloaded directly from the official databases by EleutherAI.

\textbf{Books3:} Books in the Books3 dataset were collected and processed in a variety of ways, not all of which have been publicly disclosed. Contact Shawn Presser or the Eye for further information.

\textbf{arXiv:} Papers in the arXiv dataset were uploaded by the authors, and then scraped directly from \href{https://www.arxiv.org}{arXiv} by EleutherAI.

\textbf{Github:} Code in the GitHub dataset were uploaded by the authors, and then scraped directly from \href{https://www.github.com}{GitHub} by EleutherAI.

\textbf{OpenWebText2:} Webpages in the OpenWebText2 dataset were scraped from their original webpages by EleutherAI.

\textbf{FreeLaw:} We do not know how the cases in the FreeLaw dataset were collected. Contact the Free Law Project for further details. Cases were downloaded from the Free Law Project's bulk downloader by EleutherAI.

\textbf{Wikipedia (en):} Pages in the Wikipedia (en) dataset were obtained by DeepMind from official WikiMedia dumps, uploaded to \href{https://www.kaggle.com}{Kaggle}, and then downloaded directly by EleutherAI.

\textbf{StackExchange:} Data in the StackExchange dataset were downloaded from official StackExchange dumps by EleutherAI.

\textbf{USPTO Backgrounds:} Data in the USPTO Backgrounds dataset were submitted by the patent applicants to the U.S. Patent and Trademark Office and then obtained from the U.S. Patent and Trademark Office directly.

\textbf{PubMed Abstracts:} Papers in the PubMed Abstracts dataset were uploaded by the authors, and then downloaded directly from PubMed by EleutherAI.

\textbf{Project Gutenberg (PG-19):} Books in the Project Gutenberg 1919 dataset were collected in a variety of ways, some of them not publicly known. Contact Project Gutenberg and DeepMind for more information.

\textbf{OpenSubtitles:} We do not know how the data for OpenSubtitles was collected.

\textbf{DM Mathematics:} Data in the DM Mathematics dataset were algorithmically generated by DeepMind, an artificial intelligence company and research laboratory owned by Alphabet Inc. (the parent company of Google inc.).

\textbf{BookCorpus2:} Books in the BookCorpus2 dataset were downloaded from \href{https://www.smashwords.com}{SmashWords} by EleutherAI.

\textbf{Ubuntu IRC:} Chat logs in the Ubuntu IRC dataset were archived by Ubuntu and downloaded directly by EleutherAI.

\textbf{EuroParl:} The proceedings of the European Parliament were transcribed by professional translators and transcribers, and downloaded directly by EleutherAI.

\textbf{YouTube Subtitles:} Subtitles in the YouTube Subtitles dataset were added by the video owners or volunteers, and then scraped directly from \href{https://www.youtube.com}{YouTube} by EleutherAI.

\textbf{PhilPapers:} Papers in the PhilPapers dataset were uploaded by the authors, and then downloaded directly from \href{https://www.philpapers.com}{PhilPapers} by EleutherAI. Machine-readable entries and non-english entries were kept, but entries which could not be parsed by pdfbox were ignored.

\textbf{NIH ExPORTER:} Documents in the NIH ExPORTER dataset were uploaded by their original authors, and then downloaded directly from the ExPORTER database by EleutherAI.

\textbf{HackerNews:} Comments in the HackerNews dataset were posted by their authors and scraped from the Y Combinator API directly by EleutherAI.

\textbf{Enron Emails:} Emails in the Enron Emails dataset were collected by U.S. federal agents in the course of their investigations into the Enron fraud. The data was later collected and prepared by the CALO Project.

\subsection*{If the dataset is a sample from a larger set, what was the sampling strategy (e.g., deterministic, probabilistic with specific sampling probabilities)?}
\textbf{The Pile:} All sampled datasets were sampled deterministically. How specific datasets were sampled is detailed in \citet{pile}, with code publicly available on \href{https://github.com/EleutherAI}{GitHub}.

\subsection*{Who was involved in the data collection process (e.g., students, crowdworkers, contractors) and how were they compensated (e.g., how much were crowdworkers paid)?}
\textbf{The Pile:} We do not know how data collectors were compensated for any of the datasets we did not collect ourselves. People involved with collecting new datasets for the Pile were compensated with an invitation to be an author of \citet{pile}.

\subsection*{Over what timeframe was the data collected? Does this timeframe match the creation timeframe of the data associated with the instances (e.g., recent crawl of old news articles)? If not, please describe the timeframe in which the data associated with the instances was created.}
\textbf{The Pile:} All data in the Pile was collected prior to September 1st, 2020. All data was collected between June 1st 2020 and September 1st 2020, which for the overwhelming majority of data does not overlap with the time that the data was created at.

\textbf{Pile-CC:} The earliest date of contents in Pile-CC is unknown.

\textbf{PubMed Central:} PubMed Central began when the NIH started collating papers in 2000, though it contains literature going back to the 1800s \citep{pmc_faqs}.

\textbf{Books3:} Not all details about the creation of Books3 have been publicly disclosed.  Contact Shawn Presser or the Eye for further information.

\textbf{arXiv:} We collected all papers from when arXiv launched on August 14, 1991, up to July 1, 2020.

\textbf{Github:} We collected all code from when GitHub launched on April 10, 2008, up to July 1, 2020.

\textbf{OpenWebText2:} The earliest date of contents in OpenWebText2 is unknown.

\textbf{FreeLaw:} The Free Law project has been collecting data since its founding in 2000, but some contents may be older.

\textbf{Wikipedia (en):} Wikipedia Stack Exchange launched January 15, 2001, but some contents may be older.

\textbf{StackExchange:} Stack Exchange launched in 2010, but some contents may be older.

\textbf{USPTO Backgrounds:} USPTO Backgrounds probably contains documents going back to the founding of the US Patent and Trademark Office on January 2, 1975.

\textbf{PubMed Abstracts:} This dataset took only a matter of hours to collate.

\textbf{Project Gutenberg (PG-19):} No timeframe for the collection of the PG-19 dataset is given by the authors in their paper \citep{PG19}, but the use of Project Gutenberg content would represent a recent crawl of old (pre-1919) books \citep{PG19}.

\textbf{OpenSubtitles:} We have no idea how old the contents of OpenSubtitles are.

\textbf{DM Mathematics:} DM Mathematics is a dataset of formal mathematics and therefore doesn't become ``out of date.''

\textbf{BookCorpus2:} We have no idea how old the contents of BookCorpus2 are.

\textbf{Ubuntu IRC:} No timeframe for the collection of or creation of this corpus was put forward in the relevant research paper \citep{lowe2016ubuntu}. Previous datasets created using the Ubuntu IRC put 2004-07-05 as the start date for the IRC channels themselves \citep{uthus2013ubuntu}.

\textbf{EuroParl:} The EuroParl corpus contains transcripts from 1996 to 2012.

\textbf{YouTube Subtitles:} We obtain subtitles for videos ranging from when YouTube launched on February 14, 2005 until July 1, 2020.

\textbf{PhilPapers:}  We collected the publication subset of PhilPapers which spans the years from 2009--2020.

\textbf{NIH ExPORTER:} The NIH provides a bulk-data repository for awarded applications through the ExPORTER service covering the fiscal years 1985--present. These data come from the NIH, but also other other Health and Human Services agencies (ACF, AHRQ, CDC, HRSA, FDA), and the VA. Additionally, the NIH provides a legacy data format named CRISP for awarded applications during the fiscal years 1970--2009. We merged both the ExPORTER and CRISP data to form a consolidated dataset of awarded applications.

\textbf{HackerNews:} We collect the first 24531712 posts on HackerNews. This corresponds to a date range of approximately 10/09/2006 to 09/20/2020.

\textbf{Enron Emails:} The Enron Emails were originally collected by the U.S. Department of Justice during their investigation beginning January 9, 2002 \citep{enron_doj} and took two weeks to collate in May 2002 by Joe Bartling \citep{enron_release}.

\section{Data Preprocessing}

\subsection*{Was any preprocessing/cleaning/labeling of the data done (e.g., discretization or bucketing, tokenization, part-of-speech tagging, SIFT feature extraction, removal of instances, processing of missing values)? If so, please provide a description. If not, you may skip the remainder of the questions in this section.}
\textbf{The Pile:} The data was extensively preprocessed as documented in 
\citet{pile}.

\subsection*{Was the “raw” data saved in addition to the preprocessed/cleaned/labeled data (e.g., to support unanticipated future uses)? If so, please provide a link or other access point to the “raw” data.}
\textbf{The Pile:} Yes. Access to the raw data can be obtained \href{https://github.com/EleutherAI/the-pile}{from the GitHub repo}.

\subsection*{Does this dataset collection/processing procedure achieve the motivation for creating the dataset stated in the first section of this datasheet? If not, what are the limitations?}
\textbf{The Pile:} Yes. The primary goal of the data processing is to create an extremely high quality dataset for language modeling. The Pile's success is evidenced both by its widespread adoption for training language models \citep{lieber2021jurassic,WuDao,askell2021general} and by studies of the dataset and the models trained on it such as \citet{peyrard2021invariant,mukherjee2021neural,mitchell2021fast}.

\section{Dataset Distribution}

\subsection*{How will the dataset be distributed? (e.g., tarball on website, API, GitHub; does the data have a DOI and is it archived redundantly?)}
\textbf{The Pile:} The data is distributed through several sources. It can be downloaded directly from the \href{https://github.com/EleutherAI}{EleutherAI GitHub} and from the Eye. It is archived by the Eye as well as on personal storage.

\subsection*{When will the dataset be released/first distributed? What license (if any) is it distributed under?}
\textbf{The Pile:} The dataset was released on January 1st, 2021. It is licensed under the MIT License.

\subsection*{Are there any copyrights on the data?}
\textbf{The Pile:} Some of the documents in the datasets that the Pile is based on are copyrighted. In particular, Books3 is almost entirely comprised of copyrighted works, and a substantial portion of arXiv and PhilPapers are as well. Other datasets, such as PubMed Central and GitHub contain documents that may be under limited licensing, but are not copyrighted as far as we are aware.

All data contained in the Pile has been heavily processed to aid in language modeling research and no copyrighted text is contained in the Pile in its original form. As far as we are aware, under U.S. copyright law use of copyrighted texts in the Pile falls under the ``fair use'' doctrine, which allows for the unlicensed use of copyright-protected works in some circumstances.

Copyright law varies by country, and there may be additional restrictions on some of these works in your country. If you are in doubt, it is always advisable to speak to an intellectual property attorney. If you wish to exclude some components of the Pile for legal (or any other) reason, you can compile a custom remix of the datasets using the code on the \href{https://github.com/EleutherAI}{EleutherAI GitHub} to do so.

\subsection*{Are there any fees or access/export restrictions?}

\textbf{The Pile:} There are no fees, access restrictions, or import restrictions associated with the Pile.

\section{Dataset Maintenance}

\subsection*{Who is supporting/hosting/maintaining the
dataset?}
\textbf{The Pile:} The Pile is supported and maintained by EleutherAI. The data is hosted by \href{https://the-eye.eu/}{the Eye}, and has several community maintained mirrors.

\subsection*{Will the dataset be updated? If so, how often and
by whom?}
\textbf{The Pile:} EleutherAI does not plan to update the Pile. We may release a ``Pile Version 2'' which will contain texts from a variety of languages as well as updated scrapes of the data sources that increase over time. However, in the event that such a dataset is created, it will be a separate dataset.

\subsection*{How will updates be communicated? (e.g., mailing list, GitHub)}
\textbf{The Pile:} Not applicable.

\subsection*{If the dataset becomes obsolete how will this be communicated?}
\textbf{The Pile:} If the dataset becomes obsolete, this will be communicated via our \href{https://github.com/EleutherAI}{GitHub} and \href{https://www.eleuther.ai/}{website}, as well as through various social media platforms (Twitter, Reddit, Discord).

\subsection*{Is there a repository to link to any/all papers/systems that use this dataset?}
\textbf{The Pile:} We do not maintain one, although academic databases such as Google Scholar and Semantic Scholar contain this information.

\subsection*{If others want to extend/augment/build on this dataset, is there a mechanism for them to do so? If so, is there a process for tracking/assessing the quality of those contributions. What is the process for communicating/distributing these contributions to users?}
\textbf{The Pile:} We greatly encourage community participation and feedback on the Pile. We have made all of the code necessary for constructing the Pile from scratch public to enable easier augmentation and improvement of the Pile. However we do not accept submissions of new contributions to the dataset.

\section{Legal and Ethical Considerations}

\subsection*{Were any ethical review processes conducted (e.g., by an institutional review board)? If so, please provide a description of these review processes, including the outcomes, as well as a link or other access point to any supporting documentation.}
\textbf{The Pile:} No formal ethical review process was done because EleutherAI does not have an associated IRB. Ethical considerations were discussed throughout the data collection process and is documented in our paper \citep{pile}.

\subsection*{Does the dataset contain data that might be considered confidential (e.g., data that is protected by legal privilege or by doctor-patient confidentiality, data that includes the content of individuals non-public communications)? If so, please provide a description.}

\textbf{The Pile:} We are not aware of any confidential data in the Pile, though it is always a possibility. We do know that we are not distributing any \textit{previously inaccessible} confidential data, as all data contained in the Pile was already widely and publicly available on the internet.

\subsection*{Does the dataset contain data that, if viewed directly, might be offensive, insulting, threatening, or might otherwise cause anxiety? If so, please describe why}

\textbf{The Pile:} The answer is ``probably'' for all components other than GitHub and DM Mathematics, for which the answer is ``probably not.''

\subsection*{Does the dataset identify any subpopulations (e.g., by age, gender)? If so, please describe how these subpopulations are identified and provide a description of their respective distributions within the dataset.}
\textbf{The Pile:} Answers differ by the subset. For each component document, see below.\\

\textbf{Pile-CC:} We do not know the extent to which Pile-CC identifies any subpopulations, although we expect that it does.

\textbf{PubMed Central:} Many medical papers identify subpopulations in the course of their studies. We have confirmed the presence of papers that study race, gender, ability, nation of origin, religion, and sexual orientation. 

\textbf{Books3:} Many books identify subpopulations in various ways. We have confirmed the presence of papers that study race, gender, ability, nation of origin, religion, and sexual orientation. 

\textbf{arXiv:} We do not know the extent to which arXiv identifies any subpopulations. As the overwhelming majority of papers on arXiv do not deal with human data it is most likely statistically rare. However, it is possible.

\textbf{Github:} We have no reason to believe that any GitHub code identifies any subpopulations.

\textbf{OpenWebText2:} We do not know the extent to which OpenWebText2 identifies any subpopulations, although we expect that it does.

\textbf{FreeLaw:} Many legal documents identify subpopulations in various ways. We have confirmed the presence of papers that study race, gender, ability, nation of origin, religion, and sexual orientation. 

\textbf{Wikipedia (en):} Wikipedia (en) identifies many subpopulations, and the way that this manifests and its relation to sociopolitical dynamics has been widely studied, including along the lines of race \citep{adams2019counts,xing2020editing}, gender \citep{reagle2011gender,wagner2015s,hargittai2015mind,graells2015first}, ability \citep{phillips2016wikipedia,derby2012art}, nation of origin \citep{rask2008reach,lee2017nation}, religion \citep{callahan2011cultural,ball2021using}, and sexual orientation \citep{eisner2013bi}.

\textbf{StackExchange:} We do not know the extent to which StackExchange identifies any subpopulations. As the overwhelming majority of papers on StackExchange do not deal with human data it is most likely statistically rare. However, it is possible.

\textbf{USPTO Backgrounds:} We do not know the extent to which USPTO Backgrounds identifies any subpopulations. As the overwhelming majority of abstracts submitted to the US Patent and Trademark Office do not deal with human data it is most likely statistically rare. However it is possible.

\textbf{PubMed Abstracts:} Many medical papers identify subpopulations in the course of their studies. We have confirmed the presence of papers that study race, gender, ability, nation of origin, religion, and sexual orientation. 

\textbf{Project Gutenberg (PG-19):} Many books identify subpopulations various ways. We have confirmed the presence of books that discuss race, gender, ability, nation of origin, religion, and sexual orientation. 

\textbf{OpenSubtitles:} Many movies identify subpopulations various ways. We have confirmed the presence of papers that discuss race, gender, ability, nation of origin, religion, and sexual orientation. 

\textbf{DM Mathematics:} The DM Mathematics dataset does not contain any data about people

\textbf{BookCorpus2:} Many books identify subpopulations various ways. We have confirmed the presence of papers that study race, gender, ability, nation of origin, religion, and sexual orientation. 

\textbf{Ubuntu IRC:} We do not know the extent to which Ubuntu IRC identifies any subpopulations, although we expect that it does.

\textbf{EuroParl:}  We do not know the extent to which EuroParl identifies any subpopulations, although we expect that it does.

\textbf{YouTube Subtitles:}  We do not know the extent to which YouTube Subtitles identifies any subpopulations, although we expect that it does.

\textbf{PhilPapers:} We do not know the extent to which PhilPapers identifies any subpopulations, although we expect that it does. 

\textbf{NIH ExPORTER:} Many medical papers identify subpopulations in the course of their studies. We have confirmed the presence of papers that study race, gender, ability, nation of origin, religion, and sexual orientation.

\textbf{HackerNews:} We do not know the extent to which HackerNews identifies any subpopulations, although we expect that it does.

\textbf{Enron Emails:} We do not know the extent to which Enron Emails identifies any subpopulations, although we expect that it does.

\subsection*{Is it possible to identify individuals (i.e., one or more natural persons), either directly or indirectly (i.e., in combination with other data) from the dataset? If so, please describe how.}
\textbf{The Pile:} Not applicable.

\subsection*{Does the dataset contain data that might be considered sensitive in any way (e.g., data that reveals racial or ethnic origins, sexual orientations, religious beliefs, political opinions or union memberships, or locations; financial or health data; biometric or genetic data; forms of government identification, such as social security numbers; criminal history)? If so, please provide a description.}
\textbf{The Pile:} We do not know the extent to which this is the case, although we expect that it does.

\subsection*{Did you collect the data from the individuals in question directly, or obtain it via third parties or other sources (e.g., websites)?}
\textbf{The Pile:} All data was collected from third parties, typically the original data hosts. For full details on the provenance of each dataset, see \citet{pile}.

\subsection*{Were the individuals in question notified about the data collection? If so, please describe (or show with screenshots or other information) how notice was provided, and provide a link or other access point to, or otherwise reproduce, the exact language of the notification itself.}
\textbf{The Pile:} No.

\subsection*{Did the individuals in question consent to the collection and use of their data? If so, please describe (or show with screenshots or other information) how consent was requested and provided, and provide a link or other access point to, or otherwise reproduce, the exact language to which the individuals consented.}
\textbf{The Pile:} The extent to which consent was obtained varies by dataset. For details about the provenance of each dataset, see \citet{pile}.

\subsection*{If consent was obtained, were the consenting individuals provided with a mechanism to revoke their consent in the future or for certain uses? If so, please provide a description, as well as a link or other access point to the mechanism (if appropriate).}
\textbf{The Pile:} No.

\subsection*{Has an analysis of the potential impact of the dataset and its use on data subjects (e.g., a data protection impact analysis) been conducted? If so, please provide a description of this analysis, including the outcomes, as well as a link or other access point to any supporting documentation.}
\textbf{The Pile:} No.

\bibliography{the-pile-citations,datasheet_cites}

\end{document}